\documentclass[twoside]{article}

\usepackage[accepted]{aistats2025}
\usepackage{amsthm}

\usepackage[utf8]{inputenc} 
\usepackage[T1]{fontenc}    
\usepackage{hyperref}       
\usepackage{url}            
\usepackage{booktabs}       
\usepackage{amsfonts}       
\usepackage{nicefrac}       
\usepackage{microtype}      
\usepackage[dvipsnames]{xcolor}

\usepackage{amsmath} 
\usepackage{algorithm}
\usepackage{algpseudocode}

\newtheorem{problem}{Problem}
\newtheorem{theorem}{Theorem}

\usepackage{multirow}
\usepackage{graphicx}

\def\linkpredictor{p_{\phi_1}}
\def\linkselector{p_{\phi_2}}
\def\marginal{q_{\phi_3}}
\def\h{\mathbf{h}}
\def\compgraph{\widetilde{G}}

\def\nodefeat{\mathbf{s}}
\def\edgefeat{\mathbf{e}}
\def\W{\mathbf{W}}

\newcommand{\name}{OOD-Linker}

\begin{document}

%
\runningtitle{Invariant Link Selector for Spatial-Temporal Out-of-Distribution Problem}

%

\twocolumn[

\aistatstitle{Invariant Link Selector for \\Spatial-Temporal Out-of-Distribution Problem}



\aistatsauthor{ Katherine Tieu \And Dongqi Fu \And Jun Wu \And Jingrui He }
\aistatsaddress{University of Illinois \\ Urbana-Champaign \\ \texttt{kt42@illinois.edu} \And Meta AI \\ \texttt{dongqifu@meta.com} \And Michigan State University \\ \texttt{wujun4@msu.edu} \And University of Illinois \\ Urbana-Champaign \\ \texttt{jingrui@illinois.edu}}]


\begin{abstract}
In the era of foundation models, Out-of-Distribution (OOD) problems, i.e., the data discrepancy between the training environments and testing environments, hinder AI generalization.
Further, relational data like graphs disobeying the Independent and Identically Distributed (IID) condition makes the problem more challenging, especially much harder when it is associated with time.
Motivated by this, to realize the robust invariant learning over temporal graphs, we want to investigate what components in temporal graphs are most invariant and representative with respect to labels. 
With the Information Bottleneck (IB) method, we propose an error-bounded Invariant Link Selector that can distinguish invariant components and variant components during the training process to make the deep learning model generalizable for different testing scenarios.
Besides deriving a series of rigorous generalizable optimization functions, we also equip the training with task-specific loss functions, e.g., temporal link prediction, to make pre-trained models solve real-world application tasks like citation recommendation and merchandise recommendation, as demonstrated in our experiments with state-of-the-art (SOTA) methods. Our code is available at~\url{https://github.com/kthrn22/OOD-Linker}
\end{abstract}

\section{Introduction}
Recently, foundation models have revolutionized the field of artificial intelligence, demonstrating unprecedented performance across a wide range of tasks, such as natural language processing~\cite{DBLP:journals/corr/abs-2302-13971} and computer vision~\cite{DBLP:journals/corr/abs-2307-13721}. However, these efforts face significant insufficiencies if a discrepancy between the data distributions of training and testing environments, e.g., hallucination of Large Language Models (LLMs)~\cite{DBLP:journals/corr/abs-2309-05922, DBLP:journals/csur/LiuYFJHN23, DBLP:journals/corr/abs-2408-08921}.
This cause is also referred to as Out-of-Distribution (or OOD) problem~\cite{DBLP:journals/corr/abs-2110-11334, DBLP:journals/corr/abs-2407-21794, DBLP:journals/corr/abs-2108-13624}, which mismatch poses a substantial obstacle to the generalization capabilities of AI systems, limiting their effectiveness in real-world applications.

The challenge of OOD generalization becomes even more complicated when dealing with relational data structures, particularly graphs. Unlike traditional data types, graph data inherently violates the Independent and Identically Distributed (IID) condition~\cite{DBLP:conf/aaai/0019HA23}, a fundamental assumption in many machine learning algorithms. This violation originates from the interconnected nature of graph data, where each node's features and labels are influenced by its neighbors, creating complex dependencies that are difficult to model and generalize~\cite{DBLP:conf/cikm/ZhouZF0H22, DBLP:conf/kdd/QiBH23, DBLP:conf/www/ZengZXT23, DBLP:conf/iclr/FuHXFZSWMHL24, DBLP:journals/corr/abs-2410-13798, DBLP:journals/corr/abs-2410-02296, DBLP:conf/TMLR/ZhengFMH24, DBLP:journals/corr/abs-2412-21151}.
Furthermore, the introduction of a temporal dimension to graph data exponentially increases the complexity of the problem~\cite{DBLP:conf/sigir/FuH21, DBLP:conf/aaai/YanLBJT21, DBLP:conf/kdd/FuFMTH22, DBLP:conf/www/LiFH23, DBLP:conf/nips/TieuFZHH24, DBLP:conf/nips/Lin0FQT24, DBLP:journals/corr/abs-2412-17336}. Spatial-temporal graphs, which represent evolving relationships and dynamic structures over time, present a unique set of challenges for machine learning models. The temporal aspect introduces additional variability and dependencies that must be accounted for, making the task of identifying invariant and generalizable features even more challenging.

According to the recent survey~\cite{DBLP:conf/icml/MaoCT000S0T24}, it suggests that the effective graph data required by the graph foundation models is not about the size (e.g., number of nodes and edges) but the density of different subgraph patterns; Also, there is currently no viable graph foundation models for temporal link prediction tasks, i.e., whether the link exists or not between two entities in the graph~\cite{DBLP:conf/nips/BanZLQFKTH24, DBLP:conf/icml/ZengZXZT23}. Motivated by the above analysis, our research focuses on realizing robust invariant learning over spatial-temporal graphs. The central question we aim to address is: \textbf{What components in spatial-temporal graphs are most invariant and representative with respect to labels across different domains and time periods?}

Answering this question and extracting those invariant components for utilization is crucial for developing models that can generalize effectively to unseen environments and future timestamps.
To this end, we leverage the Information Bottleneck (IB) method, a powerful framework for extracting relevant information from complex data structures. Building upon the IB principle, we propose a novel approach: an error-bounded \textbf{Invariant Link Selector}. This innovative method is designed to distinguish between invariant and variant components during the training process, enabling deep learning models to focus on the most stable and informative features of temporal graphs. Our Invariant Link Selector operates by:
\begin{itemize}
    \item Identifying and prioritizing graph components that remain consistent across different domains and time periods.
    \item Minimizing the influence of variant components that may lead to overfitting or poor generalization.
    \item Adaptively adjusting the selection process based on error bounds, ensuring robustness and reliability.
\end{itemize}

In developing this approach, we derive a series of rigorous generalizable optimization processes. These processes form the theoretical foundation of our method, providing a principled way to balance the trade-off between compressing input information and preserving relevant features for the task.

Recognizing the importance of practical applicability, we augment our method with task-specific loss functions. This integration allows our pre-trained models to be fine-tuned for real-world applications such as temporal link prediction.
To validate the effectiveness of our proposed method, we conduct extensive domain-shift experiments comparing our approach with state-of-the-art (SOTA) methods. These experiments are designed to evaluate not only the overall performance of our model but also its ability to generalize across different domains and time periods.

\section{Preliminary}
In this section, we introduce some necessary techniques and notations for paving the way for the derivation of our method.

\subsection{Information Bottleneck method}
The Information Bottleneck (IB) method~\cite{DBLP:journals/corr/physics-0004057} presents a principled approach to extract and compress the representation of relevant information from complex data structures.

Mathematically, the IB method seeks to find a compressed representation of an input variable that retains maximal information about a target variable.
Formally, given a joint distribution $p(x,y)$ over the input variable $X$ and target variable $Y$, the IB method aims to find a compressed representation $T$, which (1) compresses the input $X$ as much as possible; and (2) preserves as much relevant information as possible about the target $Y$.
Let $I(T;X)$ denote the mutual information between $T$ and $X$, and let $I(T;Y)$ denote the mutual information between $T$ and $Y$, the IB objective is to maximize:
\begin{equation}
\mathcal{L}[p(t|x)] = I(T;Y) - \beta I(T;X)
\label{eq:ib_objective}
\end{equation}
where $\beta$ is a Lagrange multiplier that controls the trade-off between compression and preservation of IB.

The optimization problem in Equation \ref{eq:ib_objective} leads to a set of self-consistent equations:
\begin{equation}
\begin{split}
p(t|x) &= K p(t) \exp(-\beta D_{KL}[p(y|x) || p(y|t)]) \label{eq:p_t_given_x} \\
p(t) &= \sum_x p(x)p(t|x), ~p(y|t) = \sum_x p(y|x)p(x|t)
\end{split}
\end{equation}
where $K$ is a scalar normalization and $D_{KL}$ is the Kullback-Leibler divergence or KL-divergence~\cite{kullback1951kullback}.

\subsection{Spatial-Temporal Graphs Modeling}
In this paper, a spatial-temporal graph can be represented by a sequence of temporal edges between two nodes, e.g., the connection between $u$ and $v$ at timestamp $t$ is $e^{t}_{u,v} =$($u$, $v$, $t$) and $t \in \{1, \ldots, T\}$.

Also, we consider the spatial-temporal graph is attributed, i.e., nodes and edges have time-evolving features, \textbf{bold vector} $\nodefeat^{t}_u$ denotes the node feature of node $u$ at time $t$, and \textbf{bold vector} $\edgefeat^{t}_{u,v}$ denotes the edge feature of the edge $e_{u,v}^{t}$ at time $t$.

For the clear notation when deriving the theoretical analysis, we denote all connections that happen at time $t$ as $G^{t}$, and we use spatial-temporal graph and temporal graph interchangeably.

\section{Proposed Method: \name}
In this section, we first formally define the problem we target to solve with concrete examples in Section~\ref{sec:problem setting}. Then, we dive into the proposed \textbf{Invariant Link Selector} with its modeling and detailed optimization procedures in Section~\ref{sec:selector}. Given that selector, in Section~\ref{sec: trainable neural architectures}, we introduce the capable neural architecture for \textbf{\name}\ to achieve pre-training and finish the link prediction in the new domain-shifted setting. The theoretical analysis of all proposed techniques is placed in Section~\ref{sec:theory}. Please refer to Appendix~\ref{app: proof-variational-bounds} for detailed proofs of variational bounds.

\subsection{Problem Setting}
\label{sec:problem setting}
Suppose we can observe the historical behavior of a temporal graph, how can we know whether a future link somewhere will appear or not, especially if we are not able to assume any condition held for the future environment, e.g., the topological structure changes, and new features and new labels emerge.

For example, in the paper citation network, after 2006, "Data Mining" papers emerge, can we use the "Information Theory" and "Database" papers interactions before 2006 to predict the citation among "Data Mining" papers?

\begin{problem}[Out-of-Distribution Generalization for Temporal Link prediction]
    For a \textbf{query link} $(u, v, T + 1)$, we need to decide whether it will happen or not at time $T+1$, i.e., label $Y_{T + 1} = 1$ means the query link exists at time $T + 1$ and $0$ vice versa.
\end{problem}

Motivated by the above question, we need to solve: \textbf{Given the unpredictable domain shift possibilities, can we disentangle the historical temporal graphs before time $T$ and identify the invariant subgraphs that are directly related to the query link and wisely use this knowledge to make the prediction?}

\subsection{Principle of Invariant Link Selector}
\label{sec:selector}
Suppose the label of a link is determined by an invariant subgraph (e.g., the cow prediction vs. camel is determined by the shape of the object, not the color of the background~\cite{DBLP:conf/nips/AhujaCZGBMR21}).

Here, we introduce how the invariant links (i.e., subgraph) are selected.
In brief, to select the most label-relevant subgraph to make predictions, our proposed selector is constrained by the mutual information between the optimal invariant subgraph and the original input graph and to remove spurious correlation towards labels, like the background color green or yellow for the cow or camel prediction. 

Hence, before involving time dimension, we can first denote the most label-relevant invariant subgraph as $G_{inv}$; then we would want to maximize the mutual information between $I(G_{inv}, Y)$ while constraining the mutual information between the original input graph $G$ and $G_{inv}$, i.e., $I(G, G_{inv}) \leq \alpha$. Mathematically, we formulate our objective as follows:
\begin{equation}
\label{eq:intractable-ib}
\arg \max_{G_{inv} \subseteq G} I(G_{inv}, Y), \text{~s.t. } I(G, G_{inv}) \leq \alpha
\end{equation}

If we can obtain a good selector function $f_{inv}$ that extracts the optimal subgraph, satisfying the constraint in Eq.~\ref{eq:intractable-ib}, then with the introduction of Lagrange multiplier $\beta$, we can re-write Eq.~\ref{eq:intractable-ib} as follows:
\begin{equation}
\label{eq:neural-network-ib}
\arg \max_{f_{inv}} I(G', Y) -\beta I(G, G'), \text{s.t. } G' = f_{inv}(G), G' \subseteq G   
\end{equation}

Then, we can involve the time. Leveraging the sequential nature of temporal graphs as we discussed in the preliminary, we re-formulate the objective as:
\begin{equation}
\label{eq:edges-ib}
\arg \max_{e_1, \dots, e_T} I(\{e\}_1^T; Y_{T + 1}) - \beta I(\{e\}_1^T; \{G\}_{1}^T)
\end{equation}
where $\{e\}_1^T = \{e^1, \dots, e^T\}, \{G\}_1^T = \{G^1, \dots, G^T\}$. For the notation clarity, $e^t$ means a bunch of selected edges that appeared at time $t$, omitting the node index in the subscript.

However, directly optimizing Eq.~\ref{eq:edges-ib} is intractable~\cite{DBLP:conf/iclr/AlemiFD017}, so we introduce approximated variational bounds that allow us to achieve Eq.~\ref{eq:edges-ib} with trainable neural architectures (details in Section~\ref{sec: trainable neural architectures}). Specifically, optimizing Eq.~\ref{eq:edges-ib} is equivalent to:
\begin{equation}
\label{eq:edge-ib-min}
\arg \min_{e_1, \dots, e_T} -I(\{e\}_1^T; Y_{T + 1}) + \beta I(\{e\}_1^T; \{G\}_{1}^T)
\end{equation}

Then, we can achieve Eq.~\ref{eq:edge-ib-min} by establishing an upper bound and minimizing this upper bound with trainable neural architectures.

Next, we elaborate on how to obtain the upper bound for each component in Eq.~\ref{eq:edge-ib-min} in the following two subsections.

\subsubsection{Minimizing $-I(\{e\}_1^T; Y_{T + 1})$}
In Eq.~\ref{eq:edge-ib-min}, we can have the first term as $-I(\{e\}_1^T; Y_{T + 1})$.

Then, it is easy to show that $-I(\{e\}_1^T; Y_{T + 1}) \leq -\log(q_{\phi_1} (Y_{T + 1} | \{e\}_{1}^{T}))$ (\textit{proof in Appendix}), where $q_{\phi_1}(Y_{T + 1} | \{e\}_1^T)$ is the variational approximation of the probability $p(Y_{T + 1} | \{e\}_1^T)$.

Thus, if we parameterize $q_{\phi_1}$ with a neural architecture, we could minimizing $-\log(q_{\phi_1} (Y_{T + 1} | \{e\}_{1}^{T}))$ by incorporating this term into the model's loss function, and thus minimizing $-I(\{e\}_1^T; Y_{T + 1})$.

\subsubsection{Minimizing $\beta I(\{e\}_1^T; \{G\}_{1}^T)$}
\label{sec: minimize-beta}
In Eq.~\ref{eq:edge-ib-min}, we can have the second term as $\beta I(\{e\}_1^T; \{G\}_{1}^T)$.

Also, we can show that $\beta I(\{e\}_1^T; \{G\}_{1}^T) \leq \beta \sum_{t = 1}^T \mathcal{D}_{KL}(p_{\phi_2}(e_t | G_t, \{e\}_{1}^{t - 1}) || q_{\phi_3}(e_t | \{e\}_1^{t - 1}))$ (\textit{proof in Appendix}), where $\mathcal{D}_{KL}$ denotes the KL-divergence.


Again, we parameterize the variational approximation of $p(e_t | G_t, \{e\}_{1}^{t - 1})$ and the prior $p(e_t | E^{t - 1})$ with $p_{\phi_2}, q_{\phi_3}$, respectively. In this way, we can integrate the upper bound into the model's loss function and minimize it, thus minimizing $\beta I(\{e\}_1^T; \{G\}_{1}^T)$.

\subsection{Selection Process and Trainable Neural Architectures}
\label{sec: trainable neural architectures}

Here, we introduce how to realize the invariant link selection (or invariant subgraph construction) through a learnable manner, such that \name\ can achieve generalizable and adaptive invariant learning and serve for effective specific tasks like temporal link prediction.

\subsubsection{Query Link and its Computational Subgraph}

To extract the invariant subgraph for a query link, we first need to define the scope for the selection. Thus, we define a computational subgraph $\compgraph_{u, v}$ for a query link $(u, v, T+1)$ as follows.

The computational graph $\compgraph_{u, v}$ should be close and thus is supposed to be the $L$-hop neighborhood graph contains links that lie within $L$-hop from edge $(u, v)$ at any time before the query time, i.e., any link within $L$-hop of edge $(u, v, t)$ in the time window, $t \in \{1, \ldots, T \}$. However, pre-defining a $L$ for all nodes across time and structure is not feasible or adaptive for the OOD setting. Therefore, we propose to adaptively learn this computational graph.

Thus, we specify the neural architecture (i.e., parameterization) for $\linkpredictor, \linkselector$, and $\marginal$. We first introduce $p_{\phi_2}, q_{\phi_3}$ for the invariant generalization and then $p_{\phi_1}$ for the task-specific label prediction, i.e., temporal link prediction task.

\subsubsection{Parameterize $p_{\phi_2}, q_{\phi_3}$}


Firstly, we model $\linkselector$ as the process of choosing a link at a certain time $t$ as an invariant link.

Formally, for a certain timestamp $t$, constructing the invariant subgraph at time $t$ can be modeled as the iterative process of modeling $p(e_t | \compgraph^t, \{e\}_1^{t - 1})$, i.e., the probability of choosing an edge as an invariant link, using neural network $\linkselector$, given the current interactions in computational graph $\compgraph^t$ and previous invariant subgraph $\{e\}_1^{t - 1}$.

In detail, for an arbitrary link $(a, b, t) \in \compgraph_{u, v}$, $\linkselector$ first maps the link to a latent representation, and then outputs the probability that $(a, b, t)$ is chosen as an invariant link.
We start by obtaining the node representation for nodes $a, b$, as Eq~\ref{eq:node-representations-agg}, by firstly aggregating information from previous invariant links and from current neighbor links, then further processing this representation with MLP transformations to obtain the probability of choosing link $(a, b, t)$ as an invariant link. Mathematically, the computational process could be described as follows.
\begin{equation}
\label{eq:node-representations-agg}
\begin{split}
\widehat{\h}^t_{a, N, \phi_2} &= \sum_{(w, t') \in \mathcal{N}(a) | t' < t} p_{(a, w, t')} [\nodefeat^{t'}_w || f_{\text{time}}(t - t') || \edgefeat^{t'}_{(w, a)}]  \\
&+ \sum_{(w, t) \in \mathcal{N}(a)} [\nodefeat^t_{w} || f_{\text{time}}(0) || \edgefeat^t_{(w, a)}] \text{,~~i.e.,}\\
\widetilde{\h}^t_{a, N, \phi_2} &= \W^{(2)}_{(agg), \phi_2}\bigg( \text{RELU} \bigg( \W^{(1)}_{(agg), \phi_2} \widehat{\h}^t_{a, N, \phi_2} \bigg) \bigg) \\
\end{split}
\end{equation}
where $[.||.]$ denotes concatenation,
$f_{\text{time}}$ is a time encoding function to obtain the vector representation of time $t$~\cite{DBLP:conf/iclr/XuRKKA20},
$\mathcal{N}(a)$ denotes the neighbors of node $a$,
$\nodefeat^{t}_w$ denotes the node feature of node $w$ at time $t$,
$\edgefeat^{t}_{(w, a)}$ denotes the edge feature of edge $(w,a)$ at time $t$,
and $p_{(a, w, t')}$ is the probability that $(a, w, t')$ is chosen as an invariant link, modeled by $\linkselector$.

After deriving the neighborhood aggregated representation for node $a$, we can obtain the node representation for $a$ as:
\begin{equation}
\label{eq:node-representations}
\begin{split}
\h^{t}_{a, N, \phi_2} = \nodefeat^t_a + \text{tanh}\bigg( \widetilde{\h}^t_{a, N, \phi_2} + \W_{\phi_2} \nodefeat^t_a \bigg)
\end{split}
\end{equation}

In the same way, we can obtain the node representation for node $b$, i.e., $\h^{t}_{b, N}$.

Next, we derive the probability that $(a, b, t)$ is chosen as an invariant link by further applying MLP transformations on the node representations $\h^t_{a, N}$ and $\h^t_{b, N}$:
\begin{equation}
\label{eq:logits-invariant-link}
\begin{split}
& \widehat{p}_{\phi_2} \big((a, b, t) \big) = \\
& \W^{(3)}_{\phi_2} \text{RELU}\bigg( \W^{(2)}_{\phi_2} \text{RELU}\bigg( \W^{(1)}_{\phi_2} [\h^t_{a, N, \phi_2} || \h^t_{b, N, \phi_2}] \bigg) \bigg) \\
\end{split}
\end{equation}
where $\W^{(3)}_{\phi_2}$, $\W^{(2)}_{\phi_2}$, and $\W^{(1)}_{\phi_2}$ are MLPs. 

After obtaining the logits $\widehat{p}(\cdot)$, we apply the Sigmoid function and, inspired by the Gumbel-Softmax parameterization trick, we incorporate the coefficient to control how "soft" the probability is. Ideally, we would want the probability to be "hard", i.e., close to $0$ or $1$, which is equivalent to choosing the link or discarding the link, as later on, we only want our model to aggregate information along invariant links. Thus, the probability is refined as follows.
\begin{equation}
\label{eq:probability-invariant-link}
\begin{split}
& \linkselector((a, b, t) \in \text{invariant subgraph}) = \\
& \text{SIGMOID}(\widehat{p}_{\phi_2}((a, b, t)) / \tau) \\
\end{split}
\end{equation}
where $\tau$ is the control coefficient, the lower $\tau$ is the "harder" the probability distribution is.

Secondly, we elaborate how we can model the prior $q(e_t | \{e\}_{t - 1}^1)$ by parameterized the distribution with a neural architecture $q_{\phi_3}$. Intuitively $q(e_t | \{e\}_{t - 1}^t)$ tells us about the probability of choosing an invariant link given we know that invariant link at previous timestamps. Therefore, similar to the parameterization $p_{\phi_2}$, we would derive node representation by neighborhood aggregation and derive a probability indicating that whether an edge contributes to the invariant subgraph or not. We formulate the computational process as follows. 
\begin{equation}
\label{eq:node-representations-agg}
\begin{split}
\widehat{\h}^t_{a, N, \phi_3} &= \sum_{(w, t') \in \mathcal{N}(a) | t' < t} p_{(a, w, t')} [\nodefeat^{t'}_w || f_{\text{time}}(t - t') || \edgefeat^{t'}_{(w, a)}] \\
\widetilde{\h}^t_{a, N, \phi_3} &= \W^{(2)}_{(agg), \phi_3}\bigg( \text{RELU} \bigg( \W^{(1)}_{(agg), \phi_3} \widehat{\h}^t_{a, N, \phi_3} \bigg) \bigg) \\
\h^{t}_{a, N, \phi_2} &= \nodefeat^t_a + \text{tanh}\bigg( \widetilde{\h}^t_{a, N, \phi_3} + \W_{\phi_3} \nodefeat^t_a \bigg)
\end{split}
\end{equation}

Similar to modeling $\linkselector$ through Eq.~\ref{eq:logits-invariant-link} and Eq.~\ref{eq:probability-invariant-link}, we obtain the probability with $\marginal$ as follows:
\begin{equation}
\label{eq:logits-invariant-link-2}
\begin{split}
& \widehat{p}_{\phi_3} \big( (a, b, t) \big) = \\
& \W^{(3)}_{\phi_2} \text{RELU}\bigg( \W^{(2)}_{\phi_2} \text{RELU}\bigg( \W^{(1)}_{\phi_1} [\h^t_{a, N, \phi_2} || \h^t_{b, N, \phi_2}] \bigg) \bigg) \\
&\text{and}\\
& \marginal((a, b, t) \in \text{invariant subgraph}) = \\
&\text{SIGMOID}(\widehat{p}_{\phi_3}((a, b, t) / \tau) \\
\end{split}
\end{equation}

\subsubsection{Parameterize $p_{\phi_1}$}

Finally, we specify the neural architecture for the link predictor, $\linkpredictor(Y_{T + 1} | \{e\}_{T}^1)$. Intuitively, we develop a neural architecture that predicts the link occurrence $(u, v, T + 1)$ based on ALL invariant links in its computational graph.

Thus, we first derive the node representation for node $u$ as follows:

\begin{equation}
\label{eq:node-representations-lp}
\begin{split}
\widehat{\h}^T_{a, N, \phi_1} &= \sum_{(w, t') \in \mathcal{N}(a) | t' < T} p_{(a, w, t')} [\nodefeat^{t'}_w || f_{\text{time}}(t - t') || \edgefeat^{t'}_{(w, a)}] \\
&+ \sum_{(w, t) \in \mathcal{N}(a)} [\nodefeat^t_{w} || f_{\text{time}}(0) || \edgefeat^t_{(w, a)}] \\
\widetilde{\h}^t_{a, N, \phi_1} &= \W^{(2)}_{(agg), \phi_1}\bigg( \text{RELU} \bigg( \W^{(1)}_{(agg), \phi_1} \widehat{\h}^t_{a, N} \bigg) \bigg) \\
\h^{t}_{a, N, \phi_1} &= \nodefeat^t_a + \text{tanh}\bigg( \widetilde{\h}^t_{a, N, \phi_1} + \W_{\phi_1} \nodefeat^t_a \bigg)
\end{split}
\end{equation}
and the link prediction is made by 
\begin{equation}
\label{eq:link-prediction}
\begin{split}
\widehat{y} = \text{sigmoid}(\W^{(2)} \text{RELU}(\W^{(1)} [(\h^{t}_u)^\top ~||~ (\h^{t}_v)^\top]^\top))
\end{split}
\end{equation}
where we also employ the sigmoid function as we are performing binary classification.
 
\subsection{Optimization of \name}

As discussed above, we first minimize the information bottleneck objective by minimizing the component's upper bound and then minimize the temporal link prediction risk.
Therefore, we can now introduce the entire loss function as follows. Given $N$ query links, we derive the loss function as:
\begin{equation}
\label{eq:loss-function}
\begin{split}
& \mathcal{L} = \frac{1}{N} \sum_{i = 1}^N -\log(p_{\phi_1}(Y_{i, T + 1} | \{e\}_{1, i}^T)) \\
&+ \sum_{i = 1}^N \sum_{t = 1}^T \mathcal{D}_{KL}(p_{\phi_2}(e_t | \compgraph^t_i, \{e\}_{1, i}^{t - 1}) || q_{\phi_3}(e_t | \{e\}_{1, i}^{t - 1}))\\   
\end{split}
\end{equation}
where the first item is for temporal link prediction, and the second item is for invariant learning.

Moreover, we denote $Y_{i, T + 1}, \compgraph^{t}_{i}, \{e\}_{1, i}^{t}$ as the label, computational graph, and invariant links corresponding to the $i$-th query link, respectively.

More comprehensive algorithmic training procedures are presented in with pseudo-code in Appendix~\ref{app: pseudocode}.


\section{Theoretical Analysis}
\label{sec:theory}
In this section, we establish an upper bound for the error difference between applying \name\ on a training distribution $\mu$ and testing distribution $\nu$ over $\mathcal{G} \times \mathcal{Y}$, where $\mathcal{G}, \mathcal{Y}$ is the space of computational graphs and labels (i.e., we can draw the computational graph of a query link and the label indicating the link occurrence from $\mu, \nu$) as follows.

\begin{theorem}
\label{theorem:1}
    Given $N$ query links and their respective computational graphs, $\compgraph_1, \ldots, \compgraph_N$, and an $\alpha-$ Lipschitz and $\sigma-$sub-Gaussian loss function $\ell$, then with probability at least $1 - \delta > 0$, we have
\begin{equation}
\label{sec:error-bound}
\begin{split}
&\bigg | \mathbb{E}_{\mu}[\ell(f(\compgraph), Y)] - \mathbb{E}_{\nu}[\ell(f(\compgraph')), Y'] \bigg| \\
&\leq \mathcal{O}\bigg( \frac{1}{N}\sum_{i = 1}^N \sqrt{2 \sigma^2 I(\phi(\compgraph_i), \compgraph_i) + D_{KL}(\mu || \nu)} \\
&+ \sqrt{\frac{\log(1 / \delta)}{N}} \bigg)
\end{split}
\end{equation}
where $f$ is our neural architecture \name with $\phi(G_i)$ denoting the invariant subgraph extracted by \name\ (Proof in Appendix).
\end{theorem}

In Eq.~\ref{sec:error-bound}, the expectations on the left hand side are taken over all $(\compgraph, Y)$ drawn from $\mu$ and $(\compgraph', Y')$ drawn from $\nu$.
Notably, our \name\ seeks to extract a predictive subgraph, while constraining the mutual information between the invariant subgraph and the original computational graph, which is equivalent to constraining $I(\phi(\compgraph_i), \compgraph_i)$, and thus constraining the error difference between two distributions. Detailed proof for Theorem~\ref{theorem:1} can be founded in Appendix~\ref{app: proof-theorem-1}.

Moreover, complexity analysis for our OOD-Linker, and comparison between our and other method's time complexity are presented in Appendix.~\ref{app: complexity}.

\section{Experiments}

In this section, we present the performance of \name\ on the temporal link prediction task under several different distribution shift settings, i.e., shifts in edge attributes and shifts in node attributes; then, we conduct an ablation study to demonstrate the robustness of the Invariant Link Selector. We provide the reproducibility details like hyperparameters, and computing resources for \name\ in the Appendix~\ref{app: reproducibility}. 

\subsection{Experimental Settings} 

\paragraph{Datasets and Baselines.} 

We examine the ability of \name\ in performing temporal link prediction with $3$ classic real-world OOD datasets, COLLAB \cite{DBLP:conf/kdd/TangWSS12}, ACT \cite{DBLP:conf/kdd/KumarZL19}, and Aminer \cite{DBLP:conf/kdd/TangZYLZS08, DBLP:conf/www/SinhaSSMEHW15} (data statistics are presented in Appendix~\ref{app: dataset-stats}), and compare \name\ against a range of baselines: (1) Static GNNs, including GAE \cite{DBLP:journals/corr/KipfW16a}, VGAE \cite{DBLP:journals/corr/KipfW16a}, are GCN\cite{DBLP:conf/iclr/KipfW17}-based autoencoders for static graphs; (2) Dynamic GNNs, including GCRN \cite{DBLP:conf/iconip/SeoDVB18}, EvolveGCN \cite{DBLP:conf/aaai/ParejaDCMSKKSL20}, DySAT \cite{DBLP:conf/wsdm/SankarWGZY20}; (3) OOD Generalization methods, including IRM \cite{DBLP:conf/nips/AhujaCZGBMR21}, V-REx \cite{DBLP:conf/icml/KruegerCJ0BZPC21}, GroupDRO \cite{DBLP:journals/corr/abs-1911-08731}; (4) Dynamic Graph OOD Generalization methods, including DIDA \cite{DBLP:conf/nips/Zhang0ZLQ022}, EAGLE \cite{DBLP:conf/nips/YuanSFZJ0023}, SILD \cite{DBLP:conf/nips/ZhangWZQWXL023}, I-DIDA \cite{DBLP:journals/corr/abs-2311-14255}.

\paragraph{Empirical evaluation details.}

Our experimental environment strictly follows the standard of SOTA OOD baselines~\cite{DBLP:conf/nips/Zhang0ZLQ022, DBLP:conf/nips/YuanSFZJ0023, DBLP:conf/nips/ZhangWZQWXL023}.
We define the out-of-distribution dataset as follows. Each dataset of COLLAB and ACT has edge attributes, so the \textit{out-of-distribution testing environment} is obtained by filtering out one certain link attribute of the original testing set.
Then, the original unfiltered testing set forms an \textit{in-distribution testing environment}.
Further, for the originally given input graph, we can obtain the train/validate/test set by performing chronological splits ($10 / 1 / 5$ for COLLAB and $20 / 2 / 8$ for ACT). To be more specific, the data splits are based on the number of distinct timestamps of the graph. For example, $10 / 1 / 5$ for COLLAB indicates that we retrieve the temporal graph snapshots of the first $10$ timestamps, and the temporal graph snapshot corresponding to the $11-$th timestamp is the validation set.
Then, we train and validate the models and select the models with the best validation score for testing them on both out-of-distribution and in-distribution testing sets. We elaborate more details on obtaining the out-of-distribution data and illustrate their distribution shifts in the Appendix~\ref{app: data-preprocess}, Appendix\ref{app: node-data-preprocess}.

\subsection{Link Prediction with Edge Attribute Shift}
\label{sec:exp-link-prediction-edge-attr}
\begin{table}
\caption{Temporal Link Prediction (ROC) in the Edge OOD Setting.}
\vspace{2mm}
\label{tab:link_prediction}
\scalebox{0.9}{
\begin{tabular}{c|cc}
\hline
Dataset                     & COLLAB     & ACT      \\
\hline
GAE               & 74.04 $\pm$ 0.75 & 60.27 $\pm$ 0.41 \\
VGAE                & 74.95 $\pm$ 1.25 & 66.29 $\pm$ 1.33 \\
GCRN                & 69.72 $\pm$ 0.45 & 64.35 $\pm$ 1.24 \\
EvolveGCN           & 76.15 $\pm$ 0.91 & 63.17 $\pm$ 1.05  \\
DySAT               & 76.59 $\pm$ 0.20 & 66.55 $\pm$ 1.21  \\
IRM                  & 75.42 $\pm$ 0.87 & 69.19 $\pm$ 1.35  \\
V-REx               & 76.24 $\pm$ 0.77 & 70.15 $\pm$ 1.09  \\
GroupDRO            & 76.33 $\pm$ 0.29 & 74.35 $\pm$ 1.62  \\
DIDA                        & 81.87 $\pm$ 0.40 & 78.64 $\pm$ 0.97 \\
EAGLE                       & \underline{84.41 $\pm$ 0.87}  &\underline{82.70 $\pm$ 0.72} \\
\hline
\name\ (OURS) & \textbf{85.30 $\pm$ 0.31} & \textbf{85.98 $\pm$ 1.00}
\\ \hline
\end{tabular}}
\end{table}

\begin{table*}[ht]
\centering
\caption{Temporal Link Prediction (ROC) in the Node OOD Setting.}
\label{tab:link-prediction-synthetic}
\begin{tabular}{cccc}
\hline
Dataset                     & COLLAB ($\bar{p} = 0.4$) & COLLAB ($\bar{p} = 0.6$) & COLLAB ($\bar{p} = 0.8$) \\
\hline
GCRN                        & 70.24 $\pm$ 1.26               & 64.01 $\pm$ 0.19               & 62.19 $\pm$ 0.39               \\
IRM                         & 69.40 $\pm$ 0.09               & 63.97 $\pm$ 0.37               & 62.66 $\pm$ 0.33               \\
V-REx                       & 70.44 $\pm$ 1.08               & 63.99 $\pm$ 0.21               & 62.21 $\pm$ 0.40               \\
GroupDRO                    & 70.30 $\pm$ 1.23               & 64.05 $\pm$ 0.21               & 62.13 $\pm$ 0.35               \\
DIDA                        & 85.20 $\pm$ 0.84               & 82.89 $\pm$ 0.23               & 72.59 $\pm$ 3.31               \\
EAGLE                       & \textbf{88.32 $\pm$ 0.61}               & $\mathbf{87.29}$ $\pm$ $\mathbf{0.71}$               & \textbf{82.30 $\pm$ 0.75}               \\
SILD                        & \underline{85.95 $\pm$ 0.18}              & \underline{84.69 $\pm$ 1.18}               & 78.01 $\pm$ 0.71               \\
I-DIDA                      & 85.27 $\pm$ 0.06               & 83.00 $\pm$ 1.08               & 74.87 $\pm$ 1.59               \\
\hline
\name\ (OURS) & 85.58 $\pm$ 1.54               & 83.09 $\pm$ 1.82               & \underline{79.83 $\pm$ 1.69}  \\ \hline           
\end{tabular}
\end{table*}

In Table~\ref{tab:link_prediction}, we report the average metric score, ROC, and the standard deviation on the testing out-of-distribution dataset. We obtain the average and standard deviation by evaluating \name~on $5$ different runs. Best OOD testing ROC score is emphasize with \textbf{bold}, and the second-best is highlighted with \underline{underline}. As suggested by Table~\ref{tab:link_prediction}, \name\ achieves the best performance on the OOD testing set, and especially for the ACT dataset, \name\ yields substantial improvements, compared to the second-best baseline, EAGLE~\cite{DBLP:conf/nips/YuanSFZJ0023}. Additionally, we report the performance comparison between our~\name and other Dynamic Graph OOD Generalization methods, including DIDA \cite{DBLP:conf/nips/Zhang0ZLQ022}, EAGLE \cite{DBLP:conf/nips/YuanSFZJ0023}, and SILD \cite{DBLP:conf/nips/ZhangWZQWXL023} for Aminer in Table~\ref{tab:link_prediction_aminer} in Appendix~\ref{app:aminer}.

\subsection{Link Prediction with Node Attribute Shift}
\label{sec:exp-link-prediction-node-attr}

To obtain the Node OOD setting, we follow the same pre-processing as SOTA baselines~\cite{DBLP:conf/nips/Zhang0ZLQ022}, \cite{DBLP:conf/nips/YuanSFZJ0023}, \cite{DBLP:conf/nips/ZhangWZQWXL023}, which are modifications from COLLAB for exhibiting node features shift. More details on how to obtain the synthetic data are provided in the Appendix. 

We report the average ROC score and the standard deviation on each synthetic testing dataset in Table~\ref{tab:link-prediction-synthetic}. We highlighted the best and second-best results with \textbf{bold} and \underline{underline}, respectively. As Table~\ref{tab:link-prediction-synthetic} suggests, our model achieves competitive results under distribution shifts of node features, as we have the second-best result on COLLAB ($\bar{p} = 0.8$), and have a close gap compared to the second-best results on COLLAB ($\bar{p} = 0.4$) and COLLAB ($\bar{p} = 0.6$). 



\subsection{Ablation Study of Invariant Link Selector}

\begin{figure*}[h]
\centering
\includegraphics[scale=0.28]{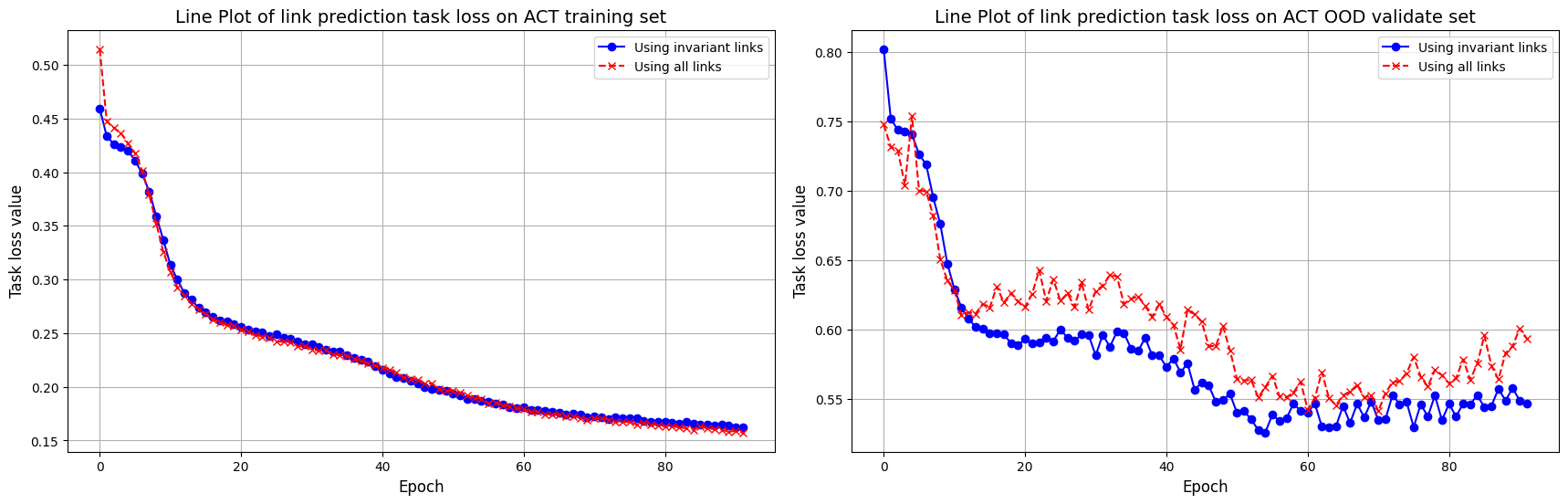}
    \caption{Comparison of link prediction task loss on the training and edge OOD settings of ACT.}
\label{fig:act_plot}
\end{figure*}

\begin{figure*}[h]
\centering
\includegraphics[scale=0.28]{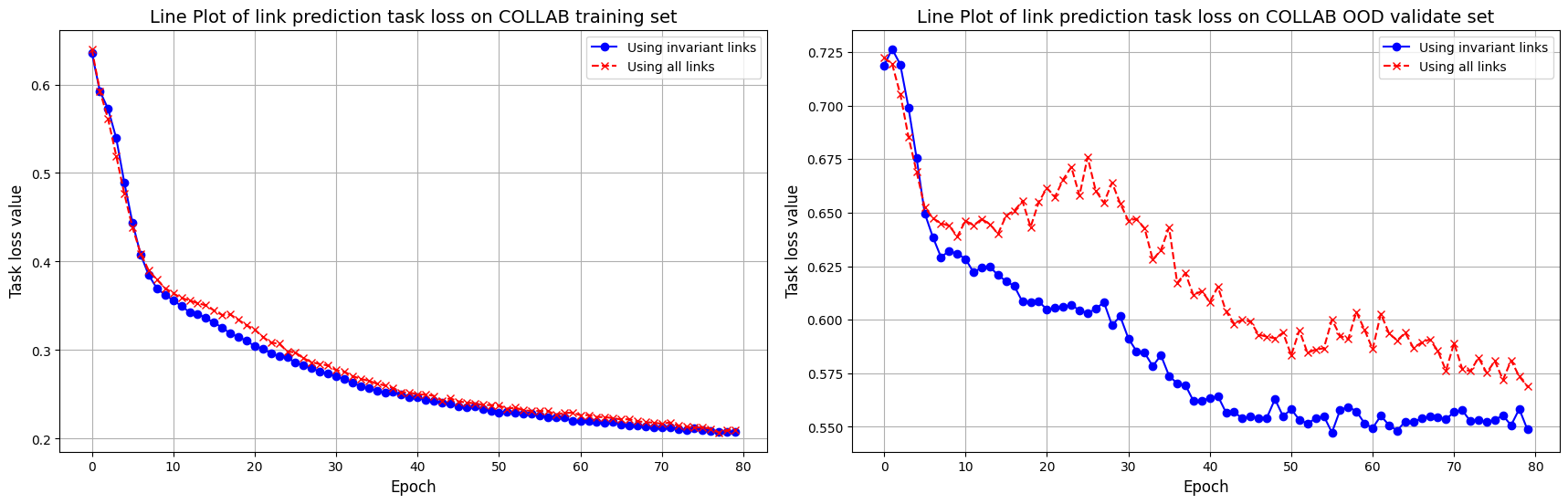}
    \caption{Comparison of link prediction task loss on the training and edge OOD settings of COLLAB.}
\label{fig:collab_plot}
\vspace{-2mm}
\end{figure*}

In this section, we provide insights into the generability of \name\ under distribution shifts by comparing the link prediction task loss values between (1) \name\ (i.e., using selected invariant links) and (2) using all links in the computational graph (i.e., omitting the invariant links discovery process).

Specifically, we present the trend of the link prediction task loss of using selected invariant links and using all links on the training and edge OOD validation sets of COLLAB and ACT in Figure~\ref{fig:act_plot} and Figure~\ref{fig:collab_plot}, respectively. In the figures, the $x$-axis represents the number of training epochs, and the $y$-axis shows the epoch-respective link prediction task loss.

As suggested by Figure~\ref{fig:act_plot} and Figure~\ref{fig:collab_plot}, the task loss of both methods on the training set decreases as the training proceeds, suggesting both models improve on prediction links in the training dataset. However, from the plot of task loss on the edge OOD validation set, we can see the gap, especially a substantial one from COLLAB, between $2$ methods. Notably, \name\ yields lower task loss on most of the epochs, suggesting that making predictions with invariant links is more robust to distribution shifts, while using all links hurts the performance in the dataset with distribution shift.

\section{Related Work}
Out-of-Distribution generalization on graphs is a critical problem in graph machine learning where model performance degrades due to distribution shifts between testing and training graph data~\cite{DBLP:journals/corr/abs-2202-07987, jiang2024graphneuralnetworkmeets, DBLP:journals/corr/abs-2402-11153, DBLP:conf/kdd/WangMWW0ZW24, DBLP:conf/kdd/ZhouHYWWZLW23, DBLP:conf/nips/XiaLWLWZZ23}.
In~\cite{DBLP:conf/nips/Zhang0ZLQ022}, the authors propose a dynamic graph neural network called DIDA to handle spatio-temporal distribution shifts in dynamic graphs. To discover variant and invariant spatio-temporal patterns in dynamic graphs, DIDA first uses a disentangled spatio-temporal attention network to encode variant and invariant patterns. It then proposes a spatio-temporal intervention mechanism to create multiple intervened distributions by sampling and reassembling variant patterns across neighborhoods and timestamps. Finally, an invariance regularization term is used to minimize the variance of predictions in the intervened distributions, enabling the model to focus on invariant patterns with stable predictive abilities to handle distribution shifts.
Motivated by DIDA~\cite{DBLP:conf/nips/Zhang0ZLQ022}, I-DIDA~\cite{DBLP:journals/corr/abs-2311-14255} is proposed. Additionally, I-DIDA further promotes the invariance property by inferring latent spatio-temporal environments and minimizing prediction variance among them.

Very recently, some new efforts have been shown to address the out-of-distribution problems in the dynamic graph deep learning community. 
A method called Spectral Invariant Learning for Dynamic Graphs under Distribution Shifts (SILD)~\cite{DBLP:conf/nips/ZhangWZQWXL023} is proposed from the spectral domain. SILD addresses two key challenges: capturing different graph patterns driven by various frequency components in the spectral domain, and handling distribution shifts with the discovered spectral patterns.
Environment-Aware Dynamic Graph Learning (EAGLE)~\cite{DBLP:conf/nips/YuanSFZJ0023} framework is also designed for out-of-distribution generalization on dynamic graphs. EAGLE addresses how to model and infer the complex environments on dynamic graphs with distribution shifts, and how to discover invariant patterns given inferred spatio-temporal environments.
EAGLE contains an environment-aware graph neural network to model environments by multi-channel environments disentangling. Then EAGLE uses an environment instantiation mechanism for environment diversification with inferred distributions. Finally, EAGLE performs fine-grained causal interventions node-wisely with a mixture of instantiated environment samples to generalize to diverse environments.
Different from previous methods on Dynamic Graphs OOD Generalization, our \name\ effectively extracts the representations that are robust to distribution shifts by looking back into the historical interactions, while other previous methods obtain invariant representations by causal intervention, which is prone to be computationally expensive. Moreover, we also establish a provable error bound, theoretically justifying \name's robustness under distribution shifts. To the best of our knowledge, we are the first effort to establish such a guarantee on Dynamic Graphs OOD Generalization. More detailed discussion on Invariant Learning and Information Bottleneck-based methods can be found in Appendix~\ref{app: detailed-related-works}.

\section{Conclusion}
In this work, we aim to advance the field of temporal graph learning and contribute to the development of more robust and generalizable deep learning models capable of handling the complex, dynamic nature of real-world relational data, i.e., spatial-temporal graphs. Hence, we propose the error-bounded Invariant Link Selector technique based on the Information Bottleneck method. Moreover, we incorporate the theoretical contribution into a concrete framework \name\ for dealing with temporal link prediction tasks in domain-shift settings, evaluated by extensive verification scenarios with SOTA baseline algorithms.


\subsubsection*{Acknowledgements}

This work is supported by National Science Foundation under Award No. IIS-2117902, and the U.S. Department of Homeland Security under Grant Award Number 17STQAC00001-08-00. The views and conclusions are those of the authors and should not be interpreted as representing the official policies of the funding agencies or the government.

\bibliography{reference}
\bibliographystyle{plain}

\section*{Checklist}



 \begin{enumerate}

 \item For all models and algorithms presented, check if you include:
 \begin{enumerate}
   \item A clear description of the mathematical setting, assumptions, algorithm, and/or model. [Yes]
   \item An analysis of the properties and complexity (time, space, sample size) of any algorithm. [Yes]
   \item (Optional) Anonymized source code, with specification of all dependencies, including external libraries. [Not Applicable]
 \end{enumerate}

 \item For any theoretical claim, check if you include:
 \begin{enumerate}
   \item Statements of the full set of assumptions of all theoretical results. [Yes]
   \item Complete proofs of all theoretical results. [Yes]
   \item Clear explanations of any assumptions. [Yes]     
 \end{enumerate}

 \item For all figures and tables that present empirical results, check if you include:
 \begin{enumerate}
   \item The code, data, and instructions needed to reproduce the main experimental results (either in the supplemental material or as a URL). [Not Applicable]
   \item All the training details (e.g., data splits, hyperparameters, how they were chosen). [Yes]
         \item A clear definition of the specific measure or statistics and error bars (e.g., with respect to the random seed after running experiments multiple times). [Yes]
         \item A description of the computing infrastructure used. (e.g., type of GPUs, internal cluster, or cloud provider). [Yes]
 \end{enumerate}

 \item If you are using existing assets (e.g., code, data, models) or curating/releasing new assets, check if you include:
 \begin{enumerate}
   \item Citations of the creator If your work uses existing assets. [Yes]
   \item The license information of the assets, if applicable. [Yes]
   \item New assets either in the supplemental material or as a URL, if applicable. [Not Applicable]
   \item Information about consent from data providers/curators. [Not Applicable]
   \item Discussion of sensible content if applicable, e.g., personally identifiable information or offensive content. [Not Applicable]
 \end{enumerate}

 \item If you used crowdsourcing or conducted research with human subjects, check if you include:
 \begin{enumerate}
   \item The full text of instructions given to participants and screenshots. [Not Applicable]
   \item Descriptions of potential participant risks, with links to Institutional Review Board (IRB) approvals if applicable. [Not Applicable]
   \item The estimated hourly wage paid to participants and the total amount spent on participant compensation. [Not Applicable]
 \end{enumerate}

 \end{enumerate}

\clearpage
\onecolumn
\appendix

\section{Proofs for Variational Bounds and Derivation of Loss function}
\label{app: proof-variational-bounds}
\subsection{Minimizing $-I(\{e\}_{1}^T; Y_{T + 1})$}

For an arbitrary $t \leq T - 1$, we have

\begin{equation}
\begin{split}
I(\{e\}_1^t; Y_{T + 1}) &= I(e_{t}, \{e\}_1^{t - 1}; Y_{T + 1}) \\
&= I(e_t; Y_{T + 1} ~|~ \{e\}_1^{t - 1}) + I(\{e\}_1^{t - 1}; Y_{T + 1})
\end{split}
\end{equation}

Thus, we have

\begin{equation}
\label{eq:mi-1-1}
\begin{split}
I(\{e\}_{1}^{T}; Y_{T + 1}) &= I(e_T, \{e\}_{1}^{T - 1}; Y_{T + 1}) \\
&= I(e_{T}; Y_{T + 1} ~|~ \{e\}_{1}^{T - 1}) + I(\{e\}_{1}^{T - 1}; Y_{T + 1}) \\
&= I(e_{T}; Y_{T + 1} ~|~ \{e\}_1^{T - 1}) + I(e_{T - 1}; Y_{T + 1} ~|~ \{e\}_1^{T - 2}) + I(\{e\}_1^{T - 2}; Y_{T + 1}) \\
& = \dots \\
&= I(e_{T}; Y_{T + 1} ~|~ \{e\}_1^{T - 1}) + I(e_{T - 1}; Y_{T + 1} ~|~ \{e\}_1^{T - 2}) + \dots + I(\{e\}_1^{1}; Y_{T + 1}) \\
\end{split}
\end{equation}

Moreover, for an arbitrary $t$, we have

\begin{equation}
\begin{split}
&I(e_t; Y_{T + 1} ~|~ \{e\}_1^{t - 1}) = \iiint P(e_t, \{e\}_{1}^{t - 1}, Y_{T + 1}) \log\frac{P(e_t, Y_{T + 1} ~|~ \{e\}_1^{t - 1})}{P(e_t ~|~ \{e\}_{1}^{t - 1}) P(Y_{T + 1} ~|~ \{e\}_{1}^{t - 1})} \\
&= \iiint P(e_t, \{e\}_{1}^{t - 1}, Y_{T + 1}) \log \frac{P(Y_{T + 1} ~|~ e_{t}, \{e\}_{1}^{t - 1}) P(e_t ~|~ \{e\}_1^{t - 1})}{P(e_t ~|~ \{e\}_{1}^{t - 1}) P(Y_{T + 1} ~|~ \{e\}_{1}^{t - 1})} \\
&= \iiint P(e_t, \{e\}_{1}^{t - 1}, Y_{T + 1}) \log \frac{P(Y_{T + 1} ~|~ e_{t}, \{e\}_{1}^{t - 1})}{P(Y_{T + 1} ~|~ \{e\}_{1}^{t - 1})} \\
&= \iiint P(e_t, \{e\}_{1}^{t - 1}, Y_{T + 1}) \log\bigg( P(Y_{T + 1} ~|~ e_t, \{e\}_{1}^{t - 1}) \bigg) - \iiint P(e_t, \{e\}_{1}^{t - 1}, Y_{T + 1}) \log \bigg( P(Y_{T + 1} ~|~ \{e\}_{1}^{t - 1}) \bigg) \\
&= \iiint P(e_t, \{e\}_{1}^{t - 1}, Y_{T + 1}) \log\bigg( P(Y_{T + 1} ~|~ \{e\}_{1}^{t}) \bigg) - \iint \log \bigg( P(Y_{T + 1} ~|~ \{e\}_{1}^{t - 1}) \bigg) \bigg(\int P(e_t, \{e\}_{1}^{t - 1}, Y_{T + 1}) \bigg) \\
&= \iint P(\{e\}_{1}^{t}, Y_{T + 1}) \log\bigg( P(Y_{T + 1} ~|~ \{e\}_{1}^{t}) \bigg) - \iint P(\{e\}_{1}^{t - 1}, Y_{T + 1}) \log \bigg( P(Y_{T + 1} ~|~ \{e\}_{1}^{t - 1}) \bigg) \\
\end{split}
\end{equation}

Thus the right-hand side of Eq.~\ref{eq:mi-1-1} is equivalent to

\begin{equation}
\begin{split}
&I(e_T; Y_{T + 1} ~|~ \{e\}_1^{T - 1}) + I(e_{T - 1}; Y_{T + 1} ~|~ \{e\}_{1}^{T - 2}) + \dots + I(\{e\}_1^1; Y_{T + 1}) = \\
&= \iint P(\{e\}_{1}^{T}, Y_{T + 1}) \log \bigg( P(Y_{T + 1} ~|~ \{e\}_{1}^{T}) \bigg) - \iint P(\{e\}_{1}^{1}, Y_{T + 1}) \log \bigg( P(Y_{T + 1}) \bigg) \\
&= \iint P(\{e\}_{1}^{T}, Y_{T + 1}) \log \bigg( P(Y_{T + 1} ~|~ \{e\}_{1}^{T}) \bigg) - \int P(Y_{T + 1}) \log \bigg( P(Y_{T + 1}) \bigg) \\
&= \iint P(\{e\}_{1}^{T}, Y_{T + 1}) \log \bigg( P(Y_{T + 1} ~|~ \{e\}_{1}^{T}) \bigg) + H(Y_{T + 1}) \\
&\geq \iint P(\{e\}_{1}^{T}, Y_{T + 1}) \log \bigg( P(Y_{T + 1} ~|~ \{e\}_{1}^T) \bigg)
\end{split}
\end{equation}

where $H(.)$ is the entropy.


However, since $P(Y_{T + 1} ~|~ \{e\}_{1}^T)$ is intractable \cite{DBLP:conf/iclr/AlemiFD017}, so let $q_{\phi_1}(Y_{T + 1} ~|~ \{e\}_{1}^T)$ be a variational approximation to $p(Y_{T + 1} ~|~ \{e\}_{1}^{T + 1})$. We have 

\begin{equation}
\begin{split}
D_{KL}(p(Y_{T + 1} ~|~ \{e\}_{1}^{T}) ~||~ q_{\phi_1}(Y_{T} ~|~ \{e\}_{1}^{T} )) &\geq 0\\
\iint p(\{e\}_{1}^{T}, Y_{T + 1}) \log \bigg( p(Y_{T + 1} ~|~ \{e\}_{1}^{T}) \bigg) &\geq \iint p(\{e\}_{1}^{T}, Y_{T + 1}) \log \bigg(q_{\phi_1}(Y_{T + 1} ~|~ \{e\}_{1}^{T}) \bigg)
\end{split}
\end{equation}

Therefore, we obtain the variational lower bound for $I(\{e\}_{1}^{T}; Y_{T+ 1})$ as follows:

\begin{equation}
\begin{split}
I(\{e\}_{1}^{T}; Y_{T + 1}) &\geq \iint p(\{e\}_{1}^{T}, Y_{T + 1}) \log \bigg( p(Y_{T + 1} ~|~ \{e\}_{1}^{T}) \bigg) \\
& \geq \iint p(\{e\}_{1}^{T}, Y_{T + 1}) \log \bigg( q_{\phi_1}(Y_{T + 1} ~|~ \{e\}_{1}^{T}) \bigg)
\end{split}
\end{equation}

Thus, $-I(\{e\}_{1}^{T}; Y_{T + 1}) \leq -\iint p(\{e\}_{1}^{T}, Y_{T + 1}) \log \bigg( q_{\phi_1} (Y_{T + 1} ~|~ \{e\}_{1}^{T}) \bigg)$

Moreover, given $N$ query links and their corresponding invariant subgraphs, $\{e\}_{1, 1}^{T}, \dots, \{e\}_{1, N}^{T}$, we can approximate the distribution $p(\{e\}_{1}^{T}, Y_{T + 1})$ as $\sum_{i = 1}^{N} \frac{1}{N} \delta_{Y_{i, T + 1}} \delta_{\{e\}_{1, i}^{T}}$, where $Y_{i, T + 1}$ is the ground-truth label of the $i-$
th query link. Thus

\begin{equation}
\label{eq:empirical}
\iint p(\{e\}_{1}^{T}, Y_{T + 1}) \log \bigg( q_{\phi_1}(Y_{T + 1} ~|~ \{e\}_{1}^T) \bigg) \approx \frac{1}{N} \sum_{i = 1}^{N} \log \bigg( q_{\phi_1}(Y_{i, T + 1} ~|~ \{e\}_{1}^T) \bigg) 
\end{equation}

Therefore, given that we parameterize $q_{\phi_1}$ with a neural architecture, we can integrate $-\frac{1}{N} \sum_{i = 1}^{N} \log \bigg( q_{\phi_1}(Y_{i, T + 1} ~|~ \{e\}_{1, i}^T) \bigg)$ into the model's loss function, and thus minimizing this term leads to minimization of $-I(\{e\}_{1}^T; Y_{T + 1})$, i.e., maximization of $I(\{e\}_{1}^{T}; Y_{T + 1})$. Moreover, we refer to $-\frac{1}{N} \sum_{i = 1}^{N} \log \bigg( q_{\phi_1}(Y_{i, T + 1} ~|~ \{e\}_{1, i}^T) \bigg)$ as the link prediction task loss, and as link prediction is a binary classification task, we employ Binary Cross Entropy to compute this component.

\subsection{Minimizing $\beta I(\{e\}_{1}^{T}; \{G\}_{1}^{T})$}

For arbitrary $t$, we have

\begin{equation}
\begin{split}
I(e_t; \{G\}_{1}^{t} ~|~ \{e\}_{1}^{t - 1}) &= I(e_{t}; G^t, \{G\}_{1}^{t - 1} ~|~ E_{1}^{t - 1}) \\
&= I(e_t; G^t ~|~ \{e\}_{1}^{t - 1}) + I(e_t; \{G\}_{1}^{t - 1} ~|~ G^t, \{e\}_{1}^{t - 1}) \\
&= I(e_t; G^t ~|~ \{e\}_{1}^{t - 1})\\
I(\{e\}_{1}^{t - 1}; \{G\}_{1}^{t}) &= I(\{e\}_{1}^{t - 1}; G^t, \{G\}_{1}^{t - 1}) \\
&= I(\{e\}_{1}^{t - 1}; \{G\}_{1}^{t - 1}) + I(\{e\}_{1}^{t - 1}; G^t ~|~ \{G\}_{1}^{t - 1}) \\
&= I(\{e\}_{1}^{t - 1}; \{G\}_{1}^{t - 1})
\end{split}
\end{equation}

As $e_{t}$ is a subset of $G^t$, so $e_t$ could be regarded as the result of a noisy function of $G^t$, i.e $e^t = f(G^t, \epsilon)$, with some noise $\epsilon$. So when $G^t$ is observed, $e^t$ becomes conditionally independent with other variables, so $I(e_t; \{G\}_{1}^{t - 1} ~|~ G^t, \{e\}_{1}^{t - 1}) = 0$. The same reasoning applies for $I(\{e\}_{1}^{t - 1}; G^t ~|~ \{G\}_{1}^{t - 1})=0$, as we can consider $\{e\}_{1}^{t - 1}$ as the result of a noisy function of $\{G\}_{1}^{t - 1}$.


Moreover, for an arbitrary $t$, we obtain the upper bound for $I(e_{t}; G^t ~|~ \{e\}_{1}^{t - 1})$ as follows. Firstly,  we have

\begin{equation}
\label{eq:mi-2-1}
\begin{split}
I(e_{t}; G^t ~|~ \{e\}_{1}^{t - 1}) &= \iiint p(e_{t}, G^t, \{e\}_{1}^{t - 1}) \log \bigg( \frac{p(e_t, G^t ~|~ \{e\}_{1}^{t - 1})}{p(e_t ~|~ \{e\}_{1}^{t - 1}) p (G^t ~|~ \{e\}_{1}^{t - 1})} \bigg) \\
&= \iiint  p(e_{t}, G^t, \{e\}_{1}^{t - 1}) \log \bigg( \frac{p(e_t ~|~ G^t, \{e\}_{1}^{t - 1}) p(G^t ~|~ \{e\}_{1}^{t - 1})}{p(e_t ~|~ \{e\}_{1}^{t - 1}) p (G^t ~|~ \{e\}_{1}^{t - 1})} \bigg) \\
&= \iiint  p(e_{t}, G^t, \{e\}_{1}^{t - 1}) \log \bigg( \frac{p(e_t ~|~ G^t, \{e\}_{1}^{t - 1})}{p(e_t ~|~ \{e\}_{1}^{t - 1})} \bigg) \\
\end{split}
\end{equation}

Let $q_{\phi_3}(e_{t} ~|~ \{e\}_{1}^{t - 1})$ be a variational approximation to $p(e_{t} ~|~ \{e\}_{1}^{t - 1})$, we have

\begin{equation}
\begin{split}
D_{KL}(p(e_t ~|~ \{e\}_{1}^{t - 1}) ~||~ q_{\phi_3}(e_t ~|~ \{e\}_{1}^{t - 1})) &\geq 0 \\
\Leftrightarrow \iint p(e_t, \{e\}_{1}^{t - 1}) \log \bigg( p(e_{t} ~|~ \{e\}_{1}^{t - 1}) \bigg) &\geq \iint p(e_t, \{e\}_{1}^{t - 1}) \log \bigg( q_{\phi_3}(e_t ~|~ \{e\}_{1}^{t - 1}) \bigg)
\end{split}
\end{equation}

Therefore, we derive the upper bound for the left-hand side of Eq.~\ref{eq:mi-2-1} as follows:

\begin{equation}
\begin{split}
&\iiint  p(e_{t}, G^t, \{e\}_{1}^{t - 1}) \log \bigg( \frac{p(e_t ~|~ G^t, \{e\}_{1}^{t - 1})}{p(e_t ~|~ \{e\}_{1}^{t - 1})} \bigg)= \\
&=\iiint p(e_{t}, G^t, \{e\}_{1}^{t - 1}) \log \bigg( p(e_t ~|~ G^t, \{e\}_{1}^{t - 1}) \bigg) - \iiint p(e_{t}, G^t, \{e\}_{1}^{t - 1}) \log \bigg( p(e_t ~|~ \{e\}_{1}^{t -  1})\bigg) \\
&= \iiint p(e_t, G^t, \{e\}_{1}^{t - 1}) \log \bigg( p(e_t ~|~ G^t, \{e\}_{1}^{t - 1}) \bigg) - \iint p(e_t, \{e\}_{1}^{t - 1}) \log \bigg( p(e_t ~|~ \{e\}_{1}^{t -  1})\bigg) \\ 
&\leq \iiint p(e_t, G^t, \{e\}_{1}^{t - 1}) \log \bigg( p(e_t ~|~ G^t, \{e\}_{1}^{t - 1}) \bigg) - \iint p(e_t, \{e\}_{1}^{t - 1}) \log \bigg( q_{\phi_3}(e_t ~|~ \{e\}_{1}^{t - 1}) \bigg)\\
&= \iiint p(e_t, G^t, \{e\}_{1}^{t - 1}) \log \bigg( p(e_t ~|~ G^t, \{e\}_{1}^{t - 1}) \bigg) - \iiint p(e_t, G^t, \{e\}_{1}^{t - 1}) \log \bigg( q_{\phi_3}(e_t ~|~ \{e\}_{1}^{t - 1}) \bigg) \\
&= \iiint p(e_t, G^t, \{e\}_{1}^{t - 1}) \log \bigg( \frac{p(e_t ~|~ G^t, \{e\}_{1}^{t - 1})}{q_{\phi_3}(e_t ~|~ \{e\}_{1}^{t - 1})} \bigg) \\
\end{split}
\end{equation}

Moreover, as we propose to parameterize $p(e_t ~|~ G^t, \{e\}_{1}^{t - 1})$ with $q_{\phi_2}(e_t ~|~ G^t, \{e\}_{1}^{t - 1})$, so we have $I(e_t; G^t ~|~ \{e\}_{1}^{t - 1}) \leq D_{KL}(q_{\phi_2}(e_t ~|~ G^t, \{e\}_{1}^{t - 1}) ~||~ q_{\phi_3}(e_t ~|~ \{e\}_{1}^{t - 1}))$. 

Putting everything together, we obtain the upper bound for $I(\{e\}_{1}^T; \{G\}_{1}^T)$ as follows:

\begin{equation}
\begin{split}
I(\{e\}_{1}^{T}; \{G\}_{1}^{T}) &= I(e_{T}; \{G\}_{1}^{T} ~|~ \{e\}_{1}^{T - 1}) + I(\{e\}_{1}^{T - 1}; \{G\}_{1}^{T}) \\
&= I(e_{T}; G^T ~|~ \{e\}_{1}^{T - 1}) + I(\{e\}_{1}^{T - 1}; \{G\}_{1}^{T - 1}) \\
&= \dots \\
&= I(e_{T}; G^T ~|~ \{e\}_{1}^{T - 1}) + I(e_{T - 1}; G^{T - 1} ~|~ \{e\}_{1}^{T - 2}) + \dots + I(e_{1}; G^1) \\
&\leq \sum_{t = 1}^T D_{KL}(q_{\phi_2}(e_t ~|~ G^t, \{e\}_{1}^{t - 1}) ~||~ q_{\phi_3}(e_t ~|~ \{e\}_{1}^{t - 1}))
\end{split}
\end{equation}

In this way, we can minimize $\beta I(\{e\}_{1}^{T}; \{G\}_{1}^{T})$ by minimizing $\beta \sum_{t = 1}^T D_{KL}(q_{\phi_2}(e_t ~|~ G^t, \{e\}_{1}^{t - 1}) ~||~ q_{\phi_3}(e_t ~|~ \{e\}_{1}^{t - 1}))$, which could be achieved by incorporating this term into the model's loss function.

\subsection{Optimization}

Given that we have obtained the variational upper bounds, we can derive the loss function for OOD-Linker's optimization process as follows. Suppose we have $N$ query links and their corresponding computational graphs $G_1, \dots, G_N$ and their respective invariant subgraphs $\{e\}_{1, 1}^{T}, \dots \{e\}_{1, N}^T$, then the loss function for our model would be

\begin{equation}
\mathcal{L} = \frac{1}{N} \sum_{i = 1}^{N} -\log \bigg( q_{\phi_1}(Y_{i, T + 1} ~|~ \{e\}_{1, i}^T) \bigg) + \beta \sum_{t = 1}^T D_{KL}(q_{\phi_2}(e_{t, i} ~|~ G^t, \{e\}_{1, i}^{t - 1}) ~||~ q_{\phi_3}(e_{t, i} ~|~ \{e\}_{1, i}^{t - 1}))
\end{equation}

Therefore, by minimizing $\mathcal{L}$, we achieve the maximization of $I(\{e\}_{1}^T; Y_{T + 1})$ and minimization of $\beta I(\{e\}_{1}^T; \{G\}_{1}^{T})$.

\section{Proof for Error Bound}
\label{app: proof-theorem-1}
In this section, we present the proof for the error bound stated in Theorem~\ref{theorem:1} in the main paper.

\begin{proof}
We start by decomposing the error difference into $2$ terms and present the proof for bounding these terms as follows.

\begin{equation}
\begin{split}
&\bigg| \mathbb{E}_{\mu}[\ell(f(\compgraph), Y)] - \mathbb{E}_{\nu}[\ell(f(\compgraph'), Y')] \bigg| =\\
&=\bigg| \mathbb{E}_{\mu}[\ell(f(\compgraph), Y)] - \frac{1}{N} \sum_{i = 1}^N \ell(f(\compgraph_i), Y_i) + \bigg( \frac{1}{N} \sum_{i = 1}^N \ell(f(\compgraph_i), Y_i) - \mathbb{E}_{\nu}[\ell(f(\compgraph'), Y')]\bigg)  \bigg| \\
&\leq \bigg| \mathbb{E}_{\mu}[\ell(f(\compgraph), Y)] - \frac{1}{N} \sum_{i = 1}^N \ell(f(\compgraph_i), Y_i) \bigg| + \bigg| \frac{1}{N}\sum_{i = 1}^N \ell(f(\compgraph_i), Y_i) - \mathbb{E}_{\nu}[\ell(f(\compgraph'), Y')] \bigg|
\end{split}
\end{equation}

The inequality holds due to Triangle Inequality. Next, we focus on bounding the two terms.

\begin{equation}
\begin{split}
&\bigg| \mathbb{E}_{\mu}[\ell(f(\compgraph), Y)] - \frac{1}{N} \sum_{i = 1}^N \ell(f(\compgraph_i), Y_i) \bigg|~ \text{  and  } ~\bigg| \frac{1}{N}\sum_{i = 1}^N \ell(f(\compgraph_i), Y_i) - \mathbb{E}_{\nu}[\ell(f(\compgraph'), Y')] \bigg| \\
\end{split}
\end{equation}

Firstly, we present how to upper-bound $\bigg| \mathbb{E}_{\mu}[\ell (f(\compgraph), Y)] - \frac{1}{N}\sum_{i = 1}^N \ell(f(\compgraph_i), Y_i) \bigg|$. Overall, we aim to bound this term by the Sequential Rademacher Complexity \cite{DBLP:conf/colt/KuznetsovM16, DBLP:journals/jmlr/RakhlinST15} of the function class containing functions such as $f$, and further bound this term by known variables.

We can rewrite $E_{\mu}[\ell(f(\compgraph), Y)]$ as $E_{\mu}[\ell (f(\compgraph^{T + 1}), Y_{T + 1}) | \compgraph^{1}, \dots \compgraph^{T}]$ due to the sequential nature of temporal graphs. With probability at least $1 - \delta$, we have

\begin{equation}
\begin{split}
& \bigg| E_{\mu}[\ell (f(\compgraph^{T + 1}), Y_{T + 1}) | \compgraph^{1}, \dots \compgraph^{T}] - \frac{1}{N} \sum_{i = 1}^N \ell(f(\compgraph_i, Y_i)) \bigg| \\
&= \bigg|E_{\mu}[\ell (f(\compgraph^{T + 1}), Y_{T + 1}) | \compgraph^{1}, \dots \compgraph^{T}] - \frac{1}{N} \sum_{i = 1}^N  \ell(f(\compgraph_i, Y_i)) \bigg| \\
&= \bigg| E_{\mu}[\ell (f(\compgraph^{T + 1}), Y_{T + 1}) | \compgraph^{1}, \dots \compgraph^{T}] - \frac{1}{N} \sum_{i = 1}^N  \sum_{i = 1}^{T}\ell(f(\compgraph^{t}_{i}, Y_i)) \bigg| \\
&\leq \mathcal{O} \bigg(\frac{1}{\sqrt{NT}} + M \mathfrak{R}^{\text{seq}}_T(\ell \circ \mathcal{F}) + \frac{M}{\sqrt{NT}}\sqrt{\log(1 / \delta}) \bigg) \\
& \leq \mathcal{O} \bigg( \alpha(\log T)^3 \mathfrak{R}^{seq}_{T}(\mathcal{F}) + \frac{M \sqrt{\log(1 / \delta)} + 1}{\sqrt{NT}} \bigg) \\
&\leq \mathcal{O} \bigg(\alpha R(\log T)^3 \sqrt{\frac{(\log T)^3}{T}} + \frac{M \sqrt{\log(1 / \delta)} + 1}{\sqrt{NT}} \bigg) \\
&= \mathcal{O} \bigg( \sqrt{\frac{\log(1 / \delta)}{N}} \bigg)
\end{split}
\end{equation}




where $\mathcal{F}$ is the class containing  functions such as $f$, and $M$ is the upper bound of the loss function, $R$ is the Lipschitz constant of MLPs transformations in neural architecture model $f$. The first inequality holds by applying Corollary $2$ from \cite{DBLP:journals/amai/KuznetsovM20}, the second inequality holds by as $\mathfrak{R}_{T}^{seq}(\ell \circ \mathcal{F}) \leq \mathcal{O}\bigg( \alpha (\log T)^3 \bigg) \mathfrak{R}_{T}^{seq}(\mathcal{F})$ as proven in \cite{DBLP:journals/corr/Mahdavi14}, and the third inequality holds due to $\mathfrak{R}_{T}^{seq}(\mathcal{F}) \leq \mathcal{O} \bigg( R \sqrt{\frac{(\log T)^3}{T}} \bigg)$ as proven in \cite{DBLP:journals/corr/Mahdavi14}.

Then, we obtain the upper bound for $\bigg| \frac{1}{N}\sum_{i = 1}^N \ell(f(\compgraph_i), Y_i) - \mathbb{E}_{\nu}[\ell(f(\compgraph'), Y')] \bigg|$ by directly apply results from \cite{DBLP:conf/isit/LiuY0L24} as follows:

\begin{equation}
\bigg| \frac{1}{N}\sum_{i = 1}^N \ell(f(\compgraph_i), Y_i) - \mathbb{E}_{\nu}[\ell(f(\compgraph), Y)] \bigg| \leq
\mathcal{O}\bigg( \frac{1}{N}\sum_{i = 1}^N \sqrt{2 \sigma^2 I(\phi(\compgraph_i), \compgraph_i) + D_{KL}(\mu || \nu)}\bigg)
\end{equation}

Combining the 2 upper bounds, we derive the error bound as:

\begin{equation}
\bigg| \mathbb{E}_{\mu}[\ell(f(\compgraph), Y)] - \mathbb{E}_{\nu}[\ell(f(\compgraph'), Y')] \bigg| \leq \mathcal{O}\bigg(\sqrt{\frac{\log(1 / \delta)}{N}} + \frac{1}{N}\sum_{i = 1}^N \sqrt{2 \sigma^2 I(\phi(\compgraph_i), \compgraph_i) + D_{KL}(\mu || \nu)}\bigg)
\end{equation}

\end{proof}

\section{More detailed discussion on related works}
\label{app: detailed-related-works}
Here, we provide discussion on more related works in Invariant Learning and comparison with methods leveraging Information Bottleneck (IB) for OOD Generalization on Graphs as follows.

\textbf{Graph Invariant Learning} methods aim to extract invariant graph patterns that could stabilize predictions under distribution shifts. For example, GIL~\cite{DBLP:conf/nips/LiZ0022} proposes to identify invariant subgraphs and infer environment labels for variant subgraphs through joint learning, GSAT~\cite{DBLP:conf/icml/MiaoLL22} learns task-relevant subgraphs by constraining mutual information with stochasticity using attention weights.  

Supported by Causal Inference theory, causal-based methods utilize Structural Causal Model (SCM)~\cite{10.1214/09-SS057} to identify and filter out spurious correlations. For instance, DIR~\cite{DBLP:conf/iclr/WuWZ0C22} filters out subgraphs that have spurious correlations with graph labels by leveraging a learnable mask, and then perform intervention with do-calculus to enhance model’s ability in recognizing invariant rationales. CIGA~\cite{DBLP:conf/nips/0002ZB00XL0C22} firsts identifying the causal subgraph using  informative invariant features (FIIF) or partially informative invariant features (PIIF), and then achieves OOD generalization by identifying the causal subgraph that maximally preserves the intra-class information across different training environments. 

Rooted in Disentanglement theories, disentangle-based methods aim at separating factors of variations in data, then proceed with learning representations by distinguishing invariant factors of the graph data.  For example, DisenGCN~\cite{DBLP:conf/icml/Ma0KW019} learns disentangled node representations by clustering neighbors into subspaces corresponding to distinct latent factors, allowing each channel to extract factor-specific features, and thus enhancing OOD Generalization. Most recently, OOD-GCL~\cite{DBLP:conf/icml/00010ZCZ024}, a self-supervised disentangled graph contrastive learning model that achieves OOD generalization without leveraging graph labels, proposed a tailored invariant self-supervised learning module to distinguish invariant and variant factors. 

Regarding notable \textbf{IB-based methods}, GSAT~\cite{DBLP:conf/icml/MiaoLL22} and DGIB~\cite{DBLP:conf/www/YuanSFJ024}, we elaborate on the differences between our \name~and aforementioned methods as follows. Although GSAT~\cite{DBLP:conf/icml/MiaoLL22} also extracts invariant subgraphs to obtain stable predictions under distribution shifts, GSAT~\cite{DBLP:conf/icml/MiaoLL22} does not consider temporal dependencies and sequential nature of dynamic graphs. Specifically, GSAT~\cite{DBLP:conf/icml/MiaoLL22}’s marginal distribution of the invariant subgraph is characterized by Bernoulli distribution as follows: the probability of an edge being included in the invariant subgraph is drawn from the Bernoulli distribution with some parameter $r$. On the contrary, our OOD-Linker considers the sequential nature of dynamic graphs, i.e., the occurrence of links at a certain time is affected by the link occurrences at previous timestamp. From the temporal setting, we reason that the invariant links at a certain time are also affected by the invariant links at previous timestamps. Therefore, for the minimization of $\beta I(\{e\}_1^T; \{G\}_1^T)$ (Section~\ref{sec: minimize-beta}), the marginal distribution (or prior), $q_{\theta_3}(.)$, of obtaining invariant links at time $t$, $e_t$, is a conditional probability, given the observed invariant links at previous timestamps, $\{e\}_{i}^{t - 1}$. In this way, the sequential nature of dynamic graphs could be taken into account. Moreover, the probability distribution of including invariant links, is also conditioned on previous invariant links. In summary, compared to GSAT~\cite{DBLP:conf/icml/MiaoLL22}, designed to operate on static graphs, our IB principle and derivation of variational bounds face additional challenges posed by dynamic graphs: temporal dependencies and sequential nature of dynamic graphs. The entanglement of temporal dependencies is also shown in our rigorous proof for obtaining the variational bounds in Appendix 1 (Supplementary Materials). Moreover, in terms of downstream tasks, GSAT~\cite{DBLP:conf/icml/MiaoLL22} is evaluated on Graph Classification, while our OOD-Linker is assessed on Temporal Link Prediction. 

DGIB, a work for dynamic graphs, but different from GSAT~\cite{DBLP:conf/icml/MiaoLL22} and our OOD-Linker, which is not targeting the invariant subgraphs. Specifically, DGIB~\cite{DBLP:conf/www/YuanSFJ024} maximizes the mutual information (MI) between the learned representation at the current timestamp $T + 1$ with the queried link label (i.e., $1$ for existing link occurrence, and $0$ otherwise) (the first term of Eq. 6 in DGIB~\cite{DBLP:conf/www/YuanSFJ024} paper) and constrains the MI between the learned representation at the current timestamp $T + 1$ with the (the second term of Eq. 6 in DGIB~\cite{DBLP:conf/www/YuanSFJ024} paper) raw historical data occurred before $T$. However, our OOD-Linker maximizes the MI between invariant links at all timestamps in the queried link’s computational subgraph and its label (first term in Eq.~\ref{eq:edges-ib}), while constrains the MI between these invariant links and the raw historical data (second term in Eq. 5). In summary, DGIB~\cite{DBLP:conf/www/YuanSFJ024} seeks to utilize IB principle to obtain the learned representation at a single timestamp, while OOD-Linker aims to extract invariant links occurring at multiple timestamps of the query link’s computational subgraph and entangles the temporal dependencies between invarian links occurring at different timestamps. Also, in terms of empirical evaluation, DGIB~\cite{DBLP:conf/www/YuanSFJ024}’s robustness is designed for the adversarial attacks (e.g., structure perturbation and feature perturbation), while our OOD-Linker evaluates the model’s robustness under distribution shifts (e.g., new unseen domain in the training process like the Data Mining papers’ citation pattern and features for the training set consisting of Medical Informatics, Theory, and Visualization papers).

\section{Experimental Details}
\subsection{Dataset details}
\label{app: dataset-stats}
\begin{table}[h]
\centering
\caption{Datasets Statistics}
\vspace{1mm}
\label{tab:dataset-stats}
\begin{tabular}{ccccc}
\hline
Dataset & $\#$ Nodes & $\#$ Links & $\#$ Unique Timestamps & Granularity \\
\hline
COLLAB & 23,035 & 151,790 & 16 & year \\
ACT & 20,408 & 202,339 & 30 & day \\
\hline
\end{tabular}
\end{table}

\begin{table}[h]
\centering
\caption{Datasets Statistics for in-distribution and out-of-distribution datasets}
\vspace{1mm}
\label{tab:dataset-stats-distribution-shift}
\begin{tabular}{ccccc}
\hline
Dataset & $\#$ Nodes & $\#$ Links & $\#$ Unique Timestamps & Granularity \\
\hline
COLLAB (in-distribution) & 19,806 & 136,996 & 16 & year \\

COLLAB (out-of-distribution) & 3,235 & 14,792 & 16 & year \\

\hline

ACT (in-distribution) & 15,274 & 184,190 & 30 & day \\

ACT (out-of-distribution) & 2,338 & 18,149 & 22 & day \\

\hline
\end{tabular}
\end{table}

In this section, we report the dataset statistics and illustrate the distribution shifts in $2$ real-world datasets used in our empirical evaluation: COLLAB~\cite{DBLP:conf/kdd/TangWSS12} and ACT~\cite{DBLP:conf/kdd/KumarZL19}. Specifically, we report the dataset statistics in Table~\ref{tab:dataset-stats}. In Table~\ref{tab:dataset-stats}, $\#$ Nodes, $\#$ Links, $\#$ Unique Timestamps, denote the number of nodes, number of edges, and number of unique timestamps of the dataset temporal graph. The dataset that yields the most challenging setting is COLLAB, which has the longest time span but has the coarsest temporal granularity, and more substantial difference in links' properties, compared to ACT.

\subsection{Out-of-distribution data for distribution shift in edge attributes}
\label{app: data-preprocess}
Next, we elaborate more details on how out-of-distribution data is obtained. Following previous works, DIDA~\cite{DBLP:conf/nips/Zhang0ZLQ022}, EAGLE~\cite{DBLP:conf/nips/YuanSFZJ0023}, and SILD~\cite{DBLP:conf/nips/ZhangWZQWXL023}, we obtain the out-of-distribution dataset based on the field information of edges as follows. 
\begin{itemize}
    \item COLLAB~\cite{DBLP:conf/kdd/TangWSS12} is an academic collaboration dataset consisting of papers published during the time window $1990$ - $2006$. Nodes denote authors, and edges denote co-authorship. There are $5$ possible attributes for a co-authorship, ``Data Mining'', ``Database'', ``Medical Informatics'', ``Theory'', and ``Visualization''. Edges associated with ``Data Mining'' are filtered out for representing the out-of-distribution dataset, while edges with other $4$ field information are merged together, forming the in-distribution dataset. For the input node features of the graph, Word2Vec~\cite{DBLP:conf/nips/MikolovSCCD13} is employed to construct $32$-dimensional node features based on paper abstracts.
    \item ACT~\cite{DBLP:conf/kdd/KumarZL19} is a dynamic graph demonstrating student actions on a MOOC platform in a month. Nodes represent students or targets of actions, and edges represent actions, respectively. K-Means~\cite{macqueen1967some} is leveraged to cluster the action features into $5$ categories, and a certain category is randomly selected to act as the shifted attribute. Then, each student or target is assigned the action features, and these features are further expanded to $32$-dimensional features with a linear function.
\end{itemize}

We report the dataset statistics for in-distribution and out-of-distribution datasets of COLLAB and ACT in Table~\ref{tab:dataset-stats-distribution-shift}.

\subsection{Synthetic data for distribution shift in node features}
\label{app: node-data-preprocess}
Here, we elaborate more on how to obtain synthetic data that exhibits distribution shifts in node attributes. Following previous work on OOD Generalization for Dynamic graphs, DIDA~\cite{DBLP:conf/nips/Zhang0ZLQ022}, EAGLE~\cite{DBLP:conf/nips/YuanSFZJ0023}, SILD~\cite{DBLP:conf/nips/ZhangWZQWXL023}, we interpolate node features of the COLLAB dataset as follows. Firstly, we sample $p(t) |\mathcal{E}^{t + 1}|$ (where $\mathcal{E}^{t + 1}$ is the number of links in the next timestamp) positive links and $(1 - p(t)) |\mathcal{E}^{t + 1}|$ negative links, then these links are further factorized into shifted features $X^{t} \in \mathbb{R}^{|\mathcal{V}| \times d}$, which is the node features for graph snapshot at time $t$. The sampling probability, $p(t)$, is varied with $\bar{p}$: $p(t) = \bar{p} + \sigma \cos(t)$. In this way, node features with higher $p(t)$ will have stronger spurious correlations with future graph snapshots. $\bar{p} = 0.1$ is used for the testing set, while for the training dataset, $\bar{p}$ is varied from $\{0.4, 0.6, 0.8\}$, corresponding to the $3$ datasets $\text{COLLAB} (\bar{p} = 0.4), \text{COLLAB} (\bar{p} = 0.6), \text{COLLAB} (\bar{p} = 0.8)$ in Table 2 of the main paper.


\subsection{Hyperparamter details}
\label{app: reproducibility}
For the purpose of reproducibility, we report the configuration and hyper-parameters in this section. Firstly, we present the hyper-parameters that are unchanged for all datasets:

\begin{itemize}
    \item Temperature, $\tau = 1.0$
    \item Learning rate: $0.0005$
    \item Batch size: $400$
\end{itemize}

\begin{table}[]
\centering
\caption{Hyper-parameters across all datasets}
\label{tab:hyper-param}
\begin{tabular}{cccc}
\hline
Dataset & Node features dimension &  Edge features dimension & Time Encoding dimension \\
\hline
COLLAB & 32 & 8 & 9 \\
ACT & 32 & 8 & 16 \\
COLLAB $(\bar{p} = 0.4)$ & 64 & 1 & 9 \\
COLLAB $(\bar{p} = 0.6)$ & 64 & 1 & 9 \\
COLLAB $(\bar{p} = 0.8)$ & 64 & 1 & 9 \\
\hline
\end{tabular}
\end{table}

Next, we report the detailed hyper-parameters across all datasets in Table~\ref{tab:hyper-param}. The experiments are implemented by Python and executed on a Linux machine with a single NVIDIA Tesla V100 32GB GPU. The code will be released upon the paper's publication.

\subsection{Additional real-world dataset performance comparison}
\label{app:aminer}

Here, we report additional empirical results on Aminer dataset \cite{DBLP:conf/kdd/TangZYLZS08, DBLP:conf/www/SinhaSSMEHW15} for our~\name, and other Dynamic Graph OOD Generalization methods, including DIDA \cite{DBLP:conf/nips/Zhang0ZLQ022}, EAGLE \cite{DBLP:conf/nips/YuanSFZJ0023}, and SILD \cite{DBLP:conf/nips/ZhangWZQWXL023}, in Table~\ref{tab:link_prediction_aminer}. Compared to other baselines for Dynamic Graph OOD Generalization, our~\name achieves the best ROC score on Aminer.

\begin{table*}[ht]
\centering
\caption{Temporal Link Prediction (ROC) in the Edge OOD Setting.}
\vspace{2mm}
\label{tab:link_prediction_aminer}
\begin{tabular}{c|ccc}
\hline
Dataset       & Aminer    \\
\hline
DIDA          & 94.06 $\pm$ 1.16 \\
EAGLE         & 96.59 $\pm$ 0.07 \\
SILD          & 94.45 $\pm$ 0.47 \\
\hline
\name\ (OURS) & \textbf{98.62 $\pm$ 0.06}
\\ \hline
\end{tabular}
\end{table*}

\section{Complexity Analysis}
\label{app: complexity}
In this section, we provide the runtime complexity analysis for \name. Given that we have $N$ query links, we aim to obtain link predictions with \name. For each query link, we would need to traverse all nodes in each $T$ discrete snapshots of its computational graphs to obtain temporal graph representations, resulting in $\mathcal{O}(|\mathcal{V}|T)$, where $|\mathcal{V}|$ is the number of nodes. Then, we compute the invariant edge weight for each link in each discrete snapshot, leading to additional $\mathcal{O}(\sum_{t = 1}^T|\mathcal{E}^t|) \leq \mathcal{O}(|\mathcal{E}|)$ cost, where $|\mathcal{E}^t|$ is the number of edges in a snapshot at time $t$, and $|\mathcal{E}|$ is the total number of edges in the temporal graphs. Thus, the total computational complexity for obtaining predictions for $N$ query links is $\mathcal{O}(N(|V|T + |E|))$, which scales linearly with the number of nodes and edges in the graph. 

\section{Pseudo-code}
\label{app: pseudocode}
In this section, we provide the pseudo-code to illustrate \name's training procedure.

\begin{algorithm}
\caption{Pseudo-code for \name's training procedure}
\label{algo:loss}
\begin{algorithmic}
\Require $N$ query links
\Ensure Link Predictions
\For {epoch in [$1$, \dots, number of epochs]}
\For {each query link $(u, v, T + 1)$}
\State Obtain computational graphs $\compgraph_{u, v}$
\For {each time $t$ in $[1, \dots, T]$}
\For {each link $(a, b, t) \in \compgraph^{t}_{u, v}$}

\State Compute $p_{\phi_2}((a, b, t) \in \text{invariant subgraph})$ as in Eq. 7, 8, 9, and 10 of the main paper

\State Compute $q_{\phi_3}((a, b, t) \in \text{invariant subgraph})$ as in Eq. 11 and 12 of the main paper
\EndFor
\EndFor
\State Temporal node representations for $u, v$: $\mathbf{h}^{T}_{u, N, \phi_1}, \mathbf{h}^{T}_{v, N, \phi_1} \leftarrow $ Eq. 13 in the main paper, using invariant edge weights $p_{\phi_2}((a, b, t)), \forall (a, b, t) \in \compgraph^{t}_{u, v}, \forall t$

\State Link prediction $\hat{y} \leftarrow $ Eq. 14 in the main paper
\EndFor

\State $\mathcal{L} \leftarrow $ Eq. 15 in the main paper, applying BCE loss for link predictions $\hat{y}$, and invariant edge weights computed by $\phi_2, \phi_3$: $p_{\phi_2}((a, b, t)), q_{\phi_3}((a, b, t)), \forall (a, b, t) \in \compgraph^{t}_{u, v}, \forall t$

\State Backpropagate loss and update the model's weights

\EndFor
\end{algorithmic}
\end{algorithm}

\vfill

\end{document}